\documentclass{article}
\usepackage{ijcai20}

\usepackage{times}

\usepackage{soul}
\usepackage{url}
\usepackage[hidelinks]{hyperref}
\usepackage[utf8]{inputenc}
\usepackage[small]{caption}
\usepackage{graphicx}
\usepackage{amsmath}
\usepackage{booktabs}
\graphicspath{ {./figures/} }
\usepackage{subcaption}

\usepackage{algorithm,algorithmic,amsmath}
\usepackage{amssymb}
\usepackage{amsmath}
\newcommand{\field}[1]{\mathbb{#1}}

\def \E{\field{E}}

\def \P{\field{P}}
\def \R{\field{R}}

\def\cA{\mathcal{A}}                                
                                
\def\cD{\mathcal{D}}

\def\cK{\mathcal{K}}

\def\cN{\mathcal{N}}

\def\cX{\mathcal{X}}

\DeclareMathOperator*{\argmax}{argmax}

\title{Selectively Contextual Bandits}

\author{
Claudia V. Roberts$^1$\and
Maria Dimakopoulou$^2$\and
Qifeng Qiao$^2$\and
Ashok Chandrashekhar$^2$\And
Tony Jebara$^3$\\
\affiliations
$^1$Princeton University\\
$^2$Netflix\\
$^3$Spotify\\
\emails
claudiar@princeton.edu,
mdimakopoulou@spotify.com,
qqia@netflix.com, 
ashok.chandrashekar@warnermedia.com,
jebara@cs.columbia.edu
}

\begin{document}

\maketitle

\begin{abstract}
Contextual bandits are widely used in industrial personalization systems. These online learning frameworks learn a treatment assignment policy in the presence of treatment effects that vary with the observed contextual features of the users. While personalization creates a rich user experience that reflect individual interests, there are benefits of a shared experience across a community that enable participation in the zeitgeist. Such benefits are emergent through network effects and are not captured in regret metrics  typically employed in evaluating bandits. To balance these needs, we propose a new online learning algorithm that preserves benefits of personalization while increasing the commonality in treatments across users. Our approach selectively interpolates between a contextual bandit algorithm and a context-free multi-arm bandit and leverages the contextual information for a treatment decision only if it promises significant gains. Apart from helping users of personalization systems balance their experience between the individualized and shared, simplifying the treatment assignment policy by making it selectively reliant on the context can help improve the rate of learning in some cases. We evaluate our approach in a classification setting using public datasets 
and show the benefits of the hybrid policy.
\end{abstract}

\section{Introduction}

In web services, users are often faced with a task of selecting an item from a large catalog. Examples include listening to a song from a catalog of 11 million on a music service or watching a video from 500 million user-uploaded videos on a video service. Equally challenging is the decision faced by the web service, the one hosting and presenting the choices to its users. Given limited screen real-estate, which of the 11 million songs or 500 million videos should it present to its users? Personalization has served as the de facto solution to this problem. Web services use proactive and reactive personalized recommendations to guide users to items that are relevant as well as help them discover new items they will also enjoy. From a mix of implicitly learned and explicitly collected features of each user and item, the services prune the list of available options that they present to each user and provide a personalized experience whilst doing so. While there are often competing objectives when deciding the optimal item or set of items to present to a user, overall, the goal is to satisfy the user in order to increase engagement with the service and retain the user over time. 
For this purpose, recommendation engines in industry rely on \emph{multi-armed bandit} algorithms to learn how to optimally recommend items to the users. In particular \emph{contextual multi-armed bandits}, which attempt to learn the optimal personalized recommendation for each user given user information, i.e., the context, are widely used in practice.

While personalization in web services makes the catalog more accessible to a user by reducing the burden of choice, it has the potential to isolate the user and unintentionally create ``filter bubbles''. This may happen even if the users are modeled by the service only using implicit behavioral data, since the feedback loops of online learning
can cause an increased focus on narrower interests.
This has the potential of hindering a user from participating 
in social conversation within their network on specific content, as the content that is served to different individuals in the network may differ drastically.
Additionally, as the user context gets more detailed and higher-dimensional, the model estimation of a contextual bandit becomes more challenging and the regret bounds may take longer to converge.
\cite{dimakopoulou2018estimation} have shown that all else equal, using assignment policies that are simpler (in terms of how they vary with contextual variables) in the early learning phases of the algorithm can improve the rate of learning and decrease regret. 
Finally, simpler, unpersonalized assignment rules may have other advantages as well; for example, \cite{lei2017actor} highlight the advantages of simplicity for interpretability in health applications of contextual bandits.

We design a new family of bandit algorithms that interpolate between unpersonalized and personalized recommendations.
This new family of algorithms, called \emph{selectively contextual bandits} (SCB), chooses between a contextual bandit decision and a non-contextual bandit decision in every iteration. The context is only used when the contextual decision yield a predicted reward lift higher than a parameter \begin{math}\delta\end{math}. 
The \begin{math}\delta\end{math} parameter is annealed at a rate that depends on the regret bound of the upper confidence bound (UCB) such that we optimize the regret bound to remove dependence on the dimensionality of the context. Alternatively, we can regularize the estimator of Lin-UCB towards a non-contextual setting. For instance, if we can show that having a \begin{math}\delta\end{math} unpersonalized (non-contextual policy) can incur at most \begin{math}{\cal O}(1)\end{math} regret per time step over an optimal contextual bandit. However, if we set $\delta$ to shrink at a certain rate, for instance, $\delta=1/t$ then we may still get a logarithmic regret.   

We evaluate our results on several contextual bandit data-sets such as classification based public datasets.
In the classification setting, we show that it is possible to achieve regret bounds equivalent to fully contextual baselines while reducing the number of contextual treatments in favor of non-contextual treatments.

\section{Preliminaries}

\subsection{Problem Formulation} \label{problem}
In the stochastic contextual bandit setting (see \cite{bubeck2012regret} for a survey), there is a finite set of arms $\cA = \{1, \dots, K\}$.
At time $t$, the environment produces $(x_t, r_t(1), \dots, r_t(K)) \sim \cD$, where $x$ is a $d$-dimensional context and $r_t(a)$ is the reward associated with each arm $a \in \cA$.

When the recommender policy selects arm $a_t$, the observables are $(x_t, a_t, r_t(a_t))$.
In particular, there is partial observability and only the reward $r_t(a_t)$ for the chosen arm $a_t$ is observed.
At each time $t$, the optimal assignment is the arm with the maximum expected reward and is denoted as \begin{math}a_t^* = \argmax_{a \in \cA} \E[r_t(a) | x_t]\end{math}. 

The goal of the policy is to find an assignment rule that sequentially assigns an arm to  minimize the cumulative expected regret over horizon $T$
\begin{align*}
\text{Regret}(T) = \sum_{t=1}^T \E [r(a^*_t) - r(a_t)]
\end{align*}
where the assignment rule is a function of the previous observations \begin{math}\left(x_\tau, a_\tau, r_\tau(a_\tau)\right)\end{math} for \begin{math}\tau=1,\dots,t-1\end{math} and of the new context $x_t$.

\subsection{Exploration vs. Exploitation} \label{explore}
Therefore, the decision-maker has to balance exploring arms for which there is limited knowledge in order to learn and exploiting the accumulated knowledge in order to attain higher rewards.

Two established approaches for balancing the exploration vs. exploitation trade-off in stochastic contextual bandits are the linear upper confidence bound (LinUCB) algorithm \cite{li2010contextual} and the linear Thompson sampling (LinTS) algorithm \cite{agrawal2013thompson} as well as their generalized linear (particularly logistic) counterparts \cite{chapelle2011empirical,li2017provably}.
These algorithms postulate that the expected reward of arm $a$ conditional on the context $x$ can be modeled as a linear or a generalized linear function of the context with unknown parameters $w_a$.
Then, at each time $t$, they use the historical observations \begin{math}\left\{\left(x_\tau, a_\tau, r_\tau(a_\tau)\right)\right\}_{\tau=1}^{t-1}\end{math} and regularized linear or logistic regression to form an upper confidence bound or posterior (exact or approximate) on the unknown parameters \begin{math}\theta_a\end{math} of each arm \begin{math}a \in \cA\end{math}. Finally,  the upper confidence bound or the posterior is used to balance exploration vs. exploitation when deciding the arm $a_t$ for the new context $x_t$.

A simple and popular heuristic for bandit problems is the \begin{math}\epsilon\end{math}-greedy exploration strategy \cite{sutton2018reinforcement}.
According to this strategy, at every time $t$ the decision-maker computes point estimates \begin{math}\hat{w}_a\end{math} for each arm \begin{math}a \in \cA\end{math} based on the historical observations \begin{math}\left\{\left(x_\tau, a_\tau, r_\tau(a_\tau)\right)\right\}_{\tau=1}^{t-1}\end{math} and uses these point estimates to find the arm with the highest predicted expected reward for context $x_t$.
Then, the decision-maker selects the best predicted arm with probability \begin{math}1-\epsilon\end{math} and with probability \begin{math}\epsilon\end{math} selects an arm from \begin{math}\cA\end{math} uniformly at random.

Both Thompson sampling and UCB for contextual bandits have strong regret bound guarantees, however Thompson sampling tends to perform much better in practice \cite{chapelle2011empirical,russo2018tutorial,dimakopoulou2018balanced}. 
On the other hand, $\epsilon$-greedy has sub-optimal guarantees compared to both Thompson sampling and UCB, but is popular in practice due to its simplicity and generally good performance.

\section{Related Work}

We focus on  algorithms for the stochastic contextual bandit problem with binary rewards, but all the presented algorithms can be extended to real number rewards.
We first present two well-known baselines from the literature; a contextual bandit algorithm that models the expected reward of each arm conditional on the context as a logistic function and a non-contextual bandit algorithm that does not take into account the context during the decision making.
Subsequently, we present our approach, the selectively contextual bandit, which interpolates between the contextual and the non-contextual bandit depending on the predicted benefit from taking the contextual information into account in each time period.
All three algorithms can be paired with any of the exploration schemes outlined in Section \ref{explore}.
Due to space limitations, we present the Thompson sampling version of the algorithms, which can be readily adapted to the UCB and $\epsilon$-greedy versions.

\subsection{$K$-Armed Bernoulli Bandit}
In the non-contextual formulation, the decision-maker does not take into account the context of every time period but rather tries to learn the unpersonalized, globally optimal arm while balancing the exploration vs. exploitation trade-off. 
One straightforward approach is to model this problem as a $K$-armed Bernoulli bandit with independent arms $\cA$ \cite{thompson1933likelihood}.
In this formulation, the reward of arm $a$ follows a Bernoulli distribution with mean $\theta_a$. 
It is standard to model the mean reward of arm $a$ using a Beta distribution with parameters $\alpha_a$ and $\beta_a$, since it is the conjugate distribution of the binomial distribution. 
At every time $t$, the agent draws a sample mean reward $\hat{\theta}_a \sim \text{Beta}\left(\alpha_a, \beta_a\right)$ for each arm $a \in \cA$ and selects arm $a_t = \argmax_{a \in \cA} \hat{\theta}_a$. 
Based on the observed reward $r_t(a_t)$, the decision-maker updates the posterior distribution on $\theta_a$.
Algorithm \ref{alg:K-armed} presents the approach.

\begin{algorithm}
\caption{Non-Contextual $K$-Armed Bernoulli Bandit}
\label{alg:K-armed}
\textbf{Require:} Initial $\alpha_a$ and $\beta_a$ for all $a \in A$ (default value: 1) 
  \begin{algorithmic}[1]
    \FOR{$t = 1, \dots, T$}
      \FOR{each arm $a \in A$}
        \STATE Sample $\hat{\theta}_a \sim \text{Beta}\left(\alpha_a, \beta_a\right)$
      \ENDFOR
      \STATE Select arm $ a_t = \argmax_{a \in A} \hat{\theta}_a $
      \STATE Observe reward $r_t(a_t)$, where $r_t \sim \cD(\cdot | x_t)$
      \IF {$r_t(a_t) = 1$}
        \STATE $\alpha_{a_t} = \alpha_{a_t} + 1$
      \ELSE
        \STATE $\beta_{a_t} = \beta_{a_t} + 1$
      \ENDIF
    \ENDFOR
  \end{algorithmic}
\end{algorithm}

\subsection{Generalized Linear Bandit} \label{GLM}
Linear bandits \cite{li2010contextual,agrawal2013thompson} and generalized linear bandits (particularly logistic) \cite{chapelle2011empirical,li2017provably}  are widely used in web services for the personalization of news recommendation, advertising and search. 
Generalized linear bandits (logistic regression in particular) have demonstrated stronger performance than linear bandits in many applications where rewards are binary.
In this section, we model the stochastic contextual bandit problem as a generalized linear bandit, as in \cite{chapelle2011empirical}.

The decision-maker models the expected reward of arm $a$, $\mu_a = \E[r(a) | x]$, as a logistic function of context $x$  with parameters  $\theta_a \in \R^d$,
$\mu_a = \P(r(a) = 1| x) = \sigma(\theta_a^\top x)$
where $\sigma(z) \equiv \frac{1}{1 + \exp(-z)}$ is the sigmoid function.
The posterior distribution on the parameters $\theta_a$ of each arm $a \in \cA$ is approximated by a multivariate Gaussian distribution updated via the Laplace approximation.
Specifically, the decision-maker starts with a multivariate Gaussian prior each $\theta_a$ with mean $\boldsymbol{\mu}_0 = \textbf{0} \in \mathbb{R}^{\ell m}$ and covariance matrix $\mathbf{\Sigma}_0 = \lambda \cdot \mathbb{I}_{\ell m}$, where $\mathbb{I}_{\ell m}$ is the ${\ell m} \times {\ell m}$ identity matrix and $\lambda$ is a regularization parameter. 
 
 \begin{algorithm}
\caption{Generalized Linear Bandit}
\label{alg:GLM}
\textbf{Require:} Parameters of weight prior $\boldsymbol{\mu}_0$ and $\boldsymbol{\Sigma}_0$ 
  \begin{algorithmic}[1]
  \STATE Draw weight sample $\hat{\textbf{w}} \sim \cN(\boldsymbol{\mu}_0, \boldsymbol{\Sigma}_0)$
    \FOR{$t = 1, \dots, T$}
      \FOR{each arm $\textbf{a} \in A$}
        \STATE Compute $\hat{\theta}(\textbf{a}) = \frac{1}{1 + \exp(- \hat{\textbf{w}}^\top {\textbf{x}_{\textbf{a}}})}$
      \ENDFOR
      \STATE Select arm $\textbf{a}_t = \argmax_\textbf{a} \hat{\theta}(\textbf{a})$
      \STATE Observe reward $r_t \sim \cD(r | \textbf{a}_t)$
      \STATE Update weight posterior parameters $\boldsymbol{\mu}_t$ and $\boldsymbol{\Sigma}_t$
      \STATE Draw a new weight sample $\hat{\textbf{w}} \sim \cN(\boldsymbol{\mu}_t, \boldsymbol{\Sigma}_t)$
    \ENDFOR
  \end{algorithmic}
\end{algorithm}

The log-posterior of $\textbf{w}$ at time $t$ is
{
\medmuskip=2mu
\thinmuskip=2mu
\thickmuskip=3mu
\begin{align*}
&\log(\mathbb{P}(\textbf{w} | \textbf{x}_{\textbf{a}_t}, r_t)) \propto
-\frac{1}{2} (\textbf{w} - \boldsymbol{\mu}_{t-1})^\top \boldsymbol{\Sigma}_{t-1}^{-1} (\textbf{w} - \boldsymbol{\mu}_{t-1}) + \\
&\qquad+r_t \log\left(  \sigma(\textbf{w}^\top {\textbf{x}_{\textbf{a}_t}}) \right) + (1- r_t) \log\left(1 -  \sigma(\textbf{w}^\top {\textbf{x}_{\textbf{a}_t}})\right) 
\end{align*}}The posterior mean of $\textbf{w}$ is the maximum a posteriori estimate $\boldsymbol{\mu}_t = \textbf{w}_{\text{MAP}} = \argmax_\textbf{w} \log(\mathbb{P}(\textbf{w} | \textbf{X}, \textbf{r})) $
and the posterior covariance matrix of $\textbf{w}$ is
$\boldsymbol{\Sigma}_t^{-1} = \boldsymbol{\Sigma}_{t-1}^{-1} +  \sigma(\textbf{w}_\text{MAP}^\top {\textbf{x}_{\textbf{a}_t}}) (1 - \sigma(\textbf{w}_\text{MAP}^\top {\textbf{x}_{\textbf{t}}})) \textbf{x}_{\textbf{a}_t} \textbf{x}_{\textbf{a}_t}^\top $.
To choose the next arm, the agent draws a weight sample $\hat{\textbf{w}} \sim \cN(\boldsymbol{\mu}_t, \boldsymbol{\Sigma}_t)$ and forms an estimate of the expected reward $\hat{\theta}(\textbf{a}) = \sigma(\hat{\textbf{w}}^\top {\textbf{x}_{\textbf{a}}})$ of each arm $\textbf{a} \in \cK$ based on this weight sample.  Then, the agent plays arm $\textbf{a}_t = \argmax_\textbf{a} \hat{\theta}(\textbf{a})$. Algorithm \ref{alg:GLM} outlines the approach.

\section{Selectively Contextual Bandit} \label{scb}
Given the constituent policies - one contextual and the other non-contextual, we now provide the details of a hybrid policy that selectively switches between the two policies. At each time step, the two policies are used to determine their optimal arm assignments. If the arms selected from the two policies are different, rewards for the two arms using the contextual policy are estimated. The estimated rewards are then compared to determine if the predicted reward from the contextual policy for the arm selected by the contextual policy is significantly better than the arm selected by the non-contextual policy. If so, SCB selects the contextual winner, if not the non-contextual winner is used. The policies are then updated with the observed reward once the SCB makes its  final selection. The overall algorithm is sketched out in Algorithm \ref{alg:scb}.
\begin{algorithm}
\caption{Selectively contextual bandits}
\label{alg:scb}
\textbf{Require:} $\pi$ : Non-contextual policy, $\pi_c( x)$ : Contextual policy
  \begin{algorithmic}[1]
    \FOR{$t = 1, \dots, T$}
      \STATE Select arm $a_{nc} = \pi$
      \STATE Select arm $a_{c} =  \pi_c(x)$
      \STATE Predict reward $r_{nc}(a_{nc})$ and $r_{c}(a_{c})$
      \IF {$\delta$($r_{c}(a_{c})$, $r_{nc}(a_{nc})$) $>$ $\lambda$}
        \STATE $a_{scb} = a_{c}$
      \ELSE
        \STATE $a_{scb} = a_{nc}$
      \ENDIF
      \STATE Update $\pi$ and $\pi_c( x)$
    \ENDFOR
  \end{algorithmic}
\end{algorithm}

In our scheme, we evaluate two different ways of calculating the non-contextual winner: mean and beta-Bernoulli. Further, we also consider two different formulations of the delta operation in the algorithm: ratio or relative difference (please find the details in Section \ref{compared-models}). Finally, we allow for shrinking or annealing the $\delta$ threshold by a constant decay rate on a specified time schedule. This is implemented by shrinking the $\delta$ threshold by a specified constant rate at specified epochs.

\section{Experiments}
We hypothesize that a fully personalized policy should always perform better in regret analysis comparisons. But there exist real-world use cases where a hybrid policy may be desired. For example, a hybrid policy may have benefits such as avoiding filter bubbles and enabling users to participate in the zeitgeist or enabling greater overlap in shared experiences in the ever increasing personalized web. No public datasets exist to verify this claim, however. So we restrict our experiments to show that the regret of a hybrid policy is comparable to a purely contextual one. We evaluate our approach in a classification setting using public datasets, and we present and discuss our results in this section. 
\subsection{Experiments on Public Datasets}
\paragraph{Multiclass Classification with Multi-Armed Bandits.} When experimenting with and comparing different contextual bandit algorithms, it is common to transform multiclass classification tasks into multi-armed bandit formulations \cite{Dudik2011}. We make the assumption that the observations are sampled from a fixed distribution and are independent and identically distributed. In both the non-contextual and contextual bandit setting, the number of classes corresponds to the number of arms. In the contextual setting, the features of each data sample correspond to that sample’s context. Accompanying each data sample is the ground truth class label. In a multiclass classification problem, the task is to learn a model that correctly assigns the correct class label to each data sample in a test set. Correspondingly, in the adaptation to a bandit problem, the goal is to learn an assignment policy that assigns the optimal (correct) arm to each sample. In the multiclass classification setting, we are attempting to learn a model that minimizes the classification error, which corresponds to the policy’s expected regret in the multi-armed bandit setting. In our implementation of the various non-contextual bandits, the assignment policy opts not to use the sample’s features (or context) during arm assignment or when updating the posterior distribution of each arm. In each non-contextual bandit’s contextual counterpart, we do leverage the additional information of the sample’s features to inform arm assignment and when updating the posterior distributions. In both the non-contextual and contextual settings, after each time time-step $t$, an arm is assigned to sample $x_t$. If the arm assignment is correct, the agent incurs a reward of one. If the arm assignment is incorrect, then the agent incurs a regret of 1. When comparing the performance of various contextual bandit algorithms in this multiclass classification problem setting, it is common to perform regret analysis and visualize the regret graphs over the history of observations using the normalized cumulative regret.
\paragraph{Experimental Set-up.} We use the Open Media Library (OpenML) \cite{OpenML2013} to collect 20 publicly available classification datasets. The datasets we use span various domains such as healthcare, biology, ecology, and computer vision and vary with regards to their attributes i.e., number of observations, classes, and features. As part of pre-processing, the categorical feature columns are one-hot encoded. Before each run, we randomly shuffle the dataset. We run our suite of bandit algorithms on each of the 20 datasets for different model hyperparameters. For the SCB bandits, we vary the various input parameters including the delta threshold value, the delta shrinkage rate, and delta shrinkage schedule. These SCB input parameters control the amount of non-contextual decisions that are made in favor of contextual ones, with the option to anneal $\delta$ by a constant rate at various time-steps in the horizon.

\paragraph{SCB Input Parameter Selection.} In this paragraph we offer a more in-depth explanation on how we pick the initial delta rates as well as how we selected the subsequent annealing rate and annealing schedule. But first, we provide some intuition for why delta rate annealing may be desired or even necessary in some cases. In the earlier timesteps of a contextual model that is learning the optimal arm assignment policy for a particular dataset, the model has not yet learned a good policy because it has not seen enough samples. Thus, we choose a higher $\delta$ value in this lower sample regime, making noncontextual decisions more frequently. As time goes on, the contextual bandit begins to learn a better arm assignment policy so we shrink the rate accordingly to account for this higher degree of confidence. During our evaluation of our SCB models on the OpenML datasets, we began by selecting an initial delta rate of 1.0 for SCB models using ratios for expected reward comparisons and 0.0 for SCB models using relative differences for expected reward comparisons. This was for sanity checking that SCB policies with thresholds meant to select the contextual winner every time matched the fully contextual baseline policy. We then chose reasonably high thresholds, e.g. 1.5 for SCB policies based on ratio comparisons and 0.5 for those based on relative difference comparisons, without an annealing schedule to observe the commutative regret over the horizon using a policy that selects a large number of noncontextual decisions in favor of contextual ones. As expected, the fully contextual bandit always performed better than our SCB models under these high initial delta rates. We then steadily decreased this initial decay rate until finding a starting rate that allowed our policy to roughly match the regret bounds of the fully contextual policy. Having found this initial delta rate, we then applied a shrinkage scheduler that was not aggressive, for doing so also decreases the number of noncontextual decisions that are made. Future work includes automatically finding the SCB parameters that allow the maximal noncontextual decisions to be made while staying on par or improving upon the fully contextual policy. 

\subsubsection{Compared Models}\label{compared-models}
The following lists the multi-armed bandit algorithms we evaluated for performance comparison. It includes different flavors of SCB bandits and baseline models. If the model name ends in ``Ratio,'' the ratio of the contextual to noncontextual expected reward is compared against the SCB delta threshold at each time-step $t$ to decide between differing arm assignments. If the model name ends in ``Diff,''  then the relative difference of the contextual and noncontextual expected reward is compared against the SCB delta threshold. $\epsilon=.2$ for all $\epsilon$-greedy bandits evaluated.
\begin{itemize}
\item \textbf{IndependentBernoulliArmsEGAgent}\cite{thompson1933likelihood} non-contextual beta-Bernoulli bandit using $\epsilon$-greedy as the explore/exploit strategy. 
\item \textbf{LogisticRegressionEGAgent}\cite{chapelle2011empirical,li2017provably} contextual bandit that models the expected reward of each arm as a logistic function using $\epsilon$-greedy as the explore/exploit strategy.
\item \textbf{SCBEGAgent\_Ratio} SCB agent that interpolates between treatment decisions made by IndependentBernoulliArmsEGAgent and LogisticRegressionEGAgent, using ratio of expected rewards for comparisons against SCB $\delta$ parameter. 
\item \textbf{meanSCBEGAgent\_Ratio} SCB agent with LogisticRegressionEGAgent as its base model, using ratio of expected rewards for comparisons against SCB $\delta$ parameter. The noncontextual winning arm is determined to be the arm with the maximum average expected reward taken across all contexts in the history. 
\item \textbf{SCBEGAgent\_Diff} SCB agent that interpolates between treatment decisions made by IndependentBernoulliArmsEGAgent and LogisticRegressionEGAgent, using the relative difference of expected rewards for comparison against SCB $\delta$ parameter.
\item \textbf{meanSCBEGAgent\_Diff} SCB agent with LogisticRegressionEGAgent as its base model, using relative differences during comparisons against SCB $\delta$ parameter. The noncontextual winning arm is determined to be the arm with the maximum average expected reward taken across all contexts in the history.
\item \textbf{IndependentBernoulliArmsTSAgent} non-contextual beta-Bernoulli bandit using Thompson Sampling as explore/exploit strategy. 
\item \textbf{LogisticRegressionTSAgent}\cite{agrawal2013thompson} contextual bandit that models the expected reward of each arm as a logistic function using Thompson Sampling as the explore/exploit strategy.
\item \textbf{SCBTSAgent\_Ratio} SCB agent that interpolates between treatment decisions made by IndependentBernoulliArmsTSAgent and LogisticRegressionTSAgent, using ratio of expected rewards for comparisons against SCB $\delta$ parameter.
\item \textbf{meanSCBTSAgent\_Ratio} SCB agent with LogisticRegressionTSAgent as its base model, using ratio of expected rewards for comparisons against SCB $\delta$ parameter. The noncontextual winning arm is determined to be the arm with the maximum average expected reward taken across all contexts in the history. 
\item \textbf{SCBTSAgent\_Diff} SCB agent that interpolates between treatment decisions made by IndependentBernoulliArmsTSAgent and LogisticRegressionTSAgent, using the relative difference of expected rewards for comparison against SCB $\delta$ parameter.
\item \textbf{meanSCBTSAgent\_Diff} SCB agent with LogisticRegressionTSAgent as its base model, using the relative difference of expected rewards for comparison against SCB $\delta$ parameter. The noncontextual winning arm is determined to be the arm with the maximum average expected reward taken across all contexts in the history.
\item \textbf{IndependentBernoulliArmsUCBAgent}\cite{li2010contextual} non-contextual beta-Bernoulli bandit using UCB as explore/exploit strategy.
\item \textbf{LogisticRegressionUCBAgent} contextual bandit that models the expected reward of each arm as a logistic function using UCB as the explore/exploit strategy.
\item \textbf{SCBUCBAgent} SCB agent that interpolates between treatment decisions made by IndependentBernoulliArmsUCBAgent and LogisticRegressionUCBAgent, using the relative difference of expected rewards for comparison against SCB $\delta$ parameter.
\item \textbf{meanSCBUCBAgent} SCB agent with LogisticRegressionUCBAgent as its base model, using the relative difference of expected rewards for comparison against SCB $\delta$ parameter. The noncontextual winning arm is determined to be the arm with the maximum average expected reward taken across all contexts in the history.
\end{itemize}

\subsubsection{Results}
\begin{figure}[ht!]
    \begin{subfigure}{0.4\textwidth}
    \includegraphics[width=\textwidth]{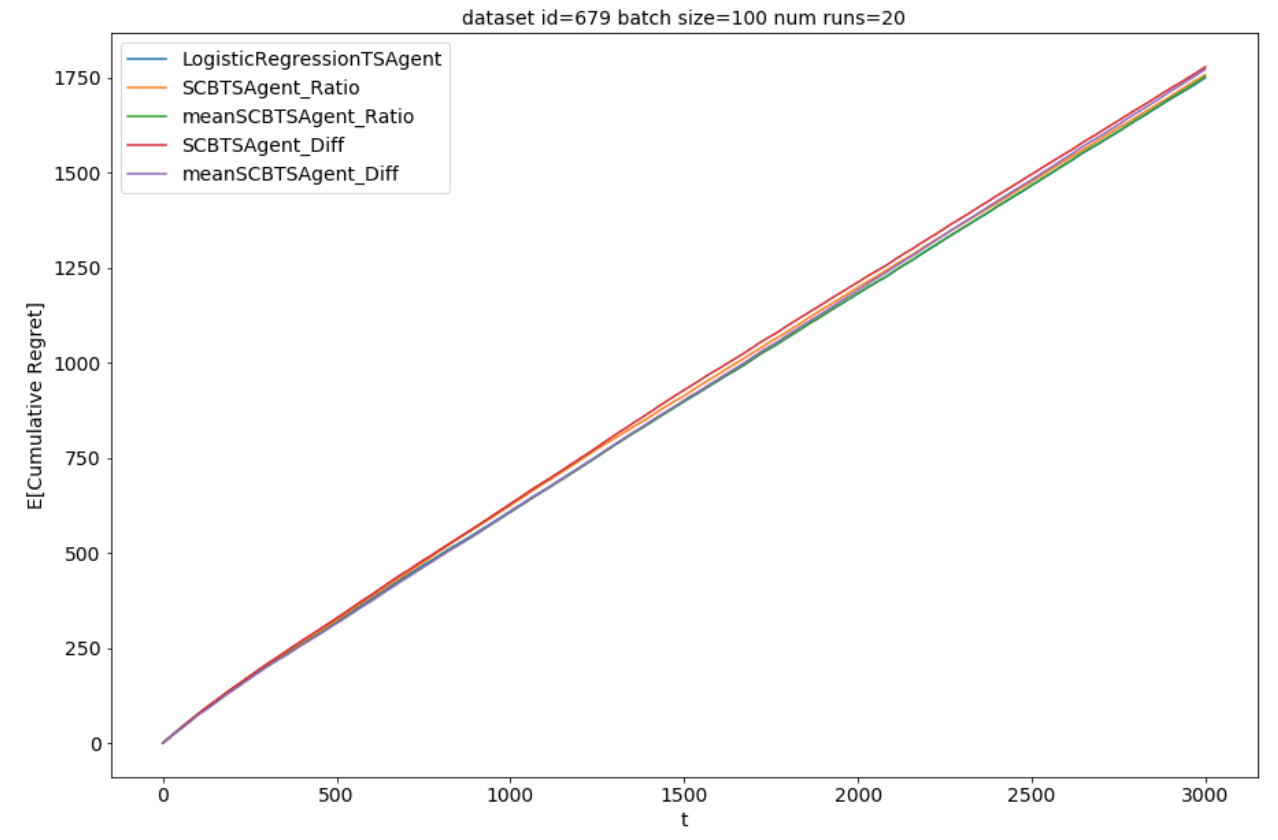}
    \caption{}
    \label{fig:679-ts}
    \end{subfigure}
    \hfill
    \begin{subfigure}{0.4\textwidth}
    \includegraphics[width=\textwidth]{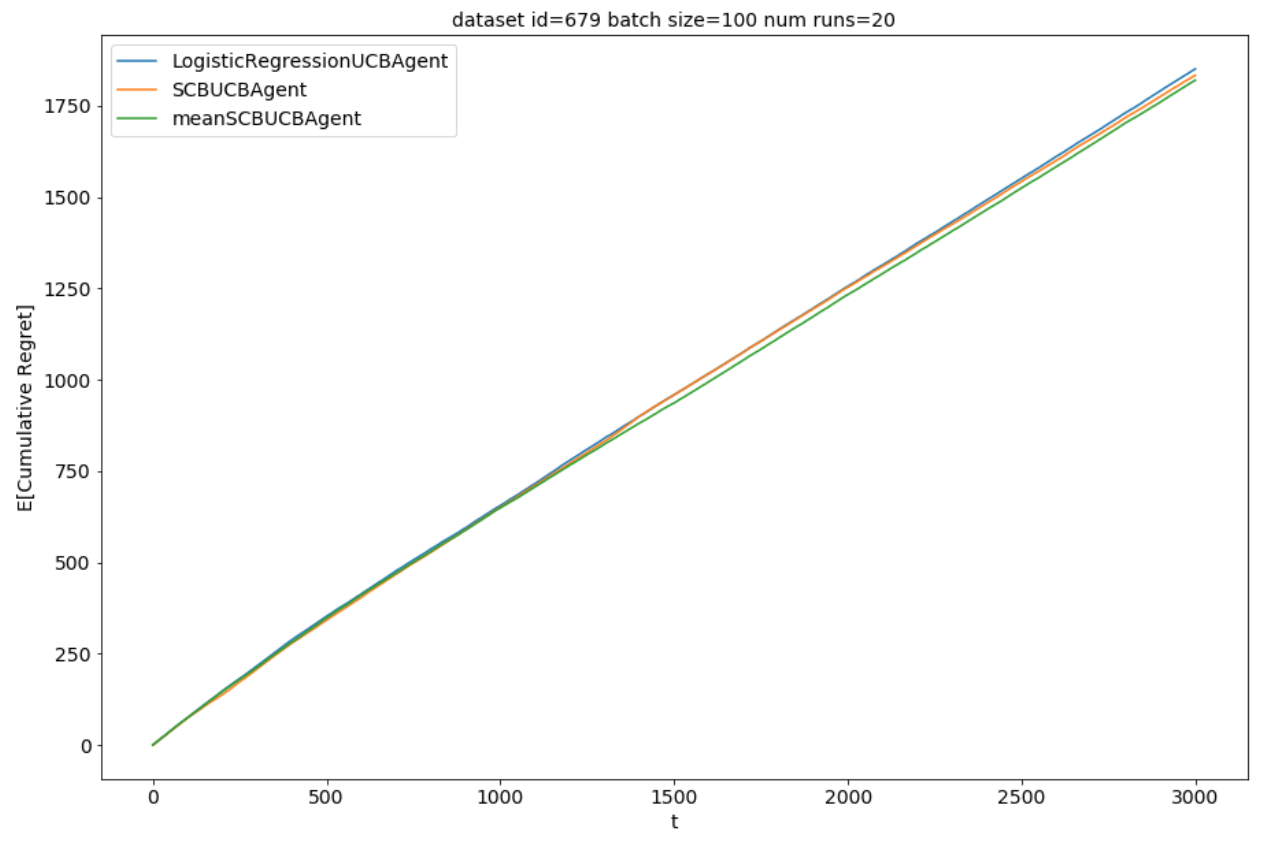}
    \caption{}
    \label{fig:679-ucb}
    \end{subfigure}
    \hfill
    \begin{subfigure}{0.4\textwidth}
    \includegraphics[width=\textwidth]{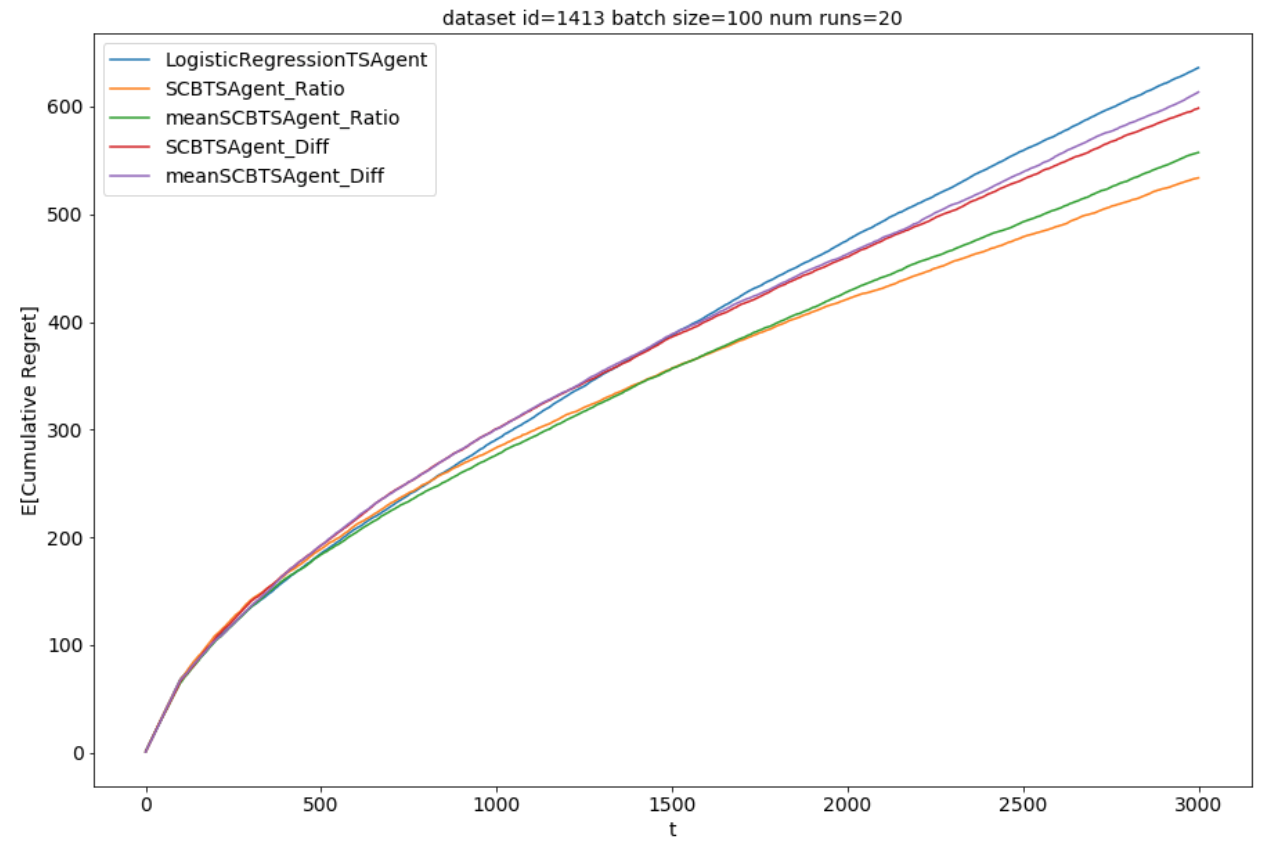}
    \caption{}
    \label{fig:1413-ts}
    \end{subfigure}
    \hfill
    \begin{subfigure}{0.4\textwidth}
    \includegraphics[width=\textwidth]{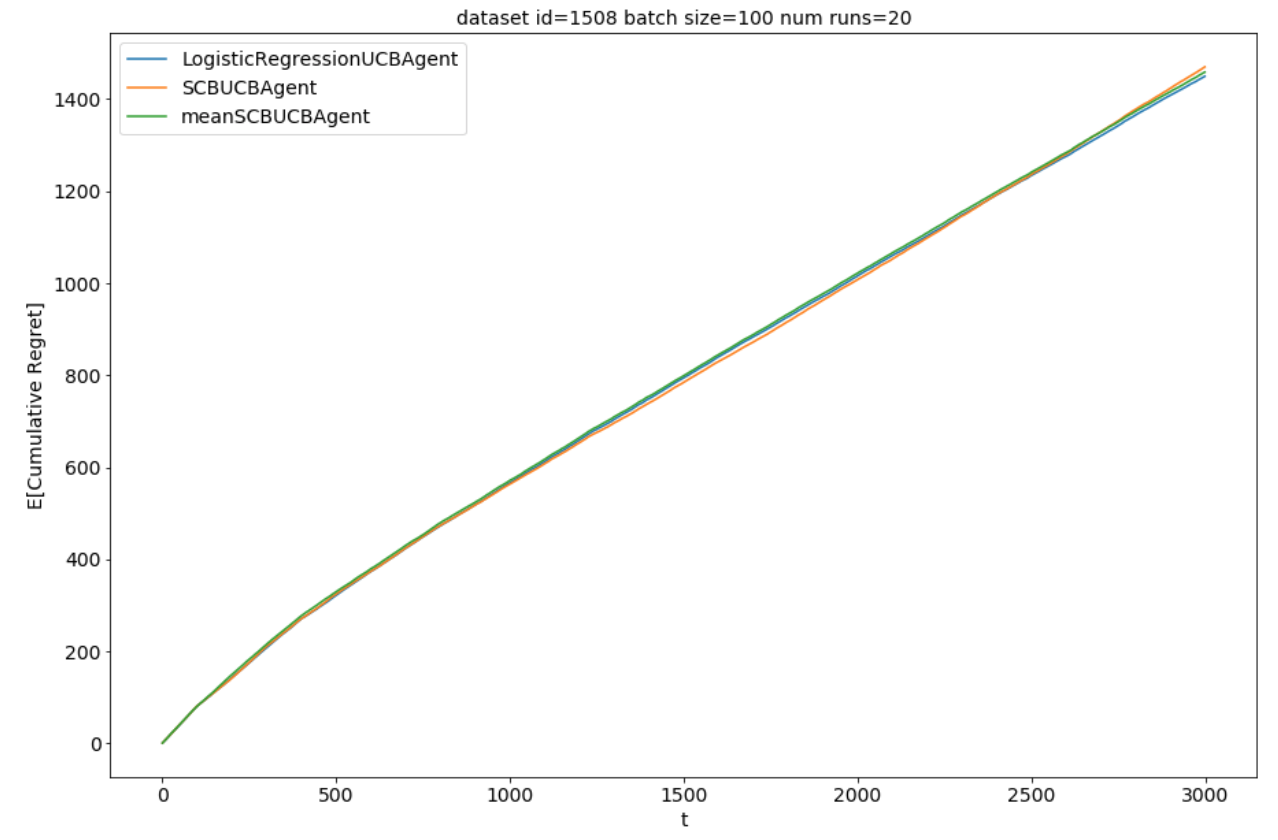}
    \caption{}
    \label{fig:1508-ucb}
    \end{subfigure}
    \caption{Regret analysis comparison between SCB and baseline algorithms on OpenML public multiclass classification datasets.}
    \label{fig:openml}
\end{figure}
Figure \ref{fig:openml} shows a regret analysis comparison between the various SCB and baseline algorithms described in Section \ref{compared-models} on three different OpenML public multiclass classification datasets. Due to space constraints, we have not included the full set of graphs for each of the 20 datasets on each of the 16 models. The $x$ coordinate is the timestep and the $y$ coordinate is the normalized cumulative regret $\frac{1}{n}\sum_{t=1}^{n} (1-r_t(a_t))$. $t$ is the timestep, the arm selected by the bandit at time $t$ is $a_t \in \{1,2,...K\}$, where $K$ is the number of arms or classes in the dataset with $n$ observations and with reward function $r_t(a_t) = 1 \{a_t = c_t\}$ where $c_t$ is the true class of context $x_t$. The lower the cumulative regret the better. We run each agent 20 times over each dataset (the dataset is reshuffled at the beginning of each run). We set the horizon to 3000 observations, updating the prior distributions and the logistic models after every batch of 100 timesteps. 

Figure \ref{fig:679-ts} shows a comparison of the various Thompson Sampling bandits we evaluated on OpenML dataset id 679, a dataset of sleep state measurements. We observe that all five bandits display the same regret curves. However, while the baseline model, LogisticRegressionTSAgent, made contextual decisions at every timestep $t$, the SCBTSAgent\_Ratio bandit chose noncontextual decisions over contextual ones 7.2\% of the time averaged across the 20 runs. Similarly, the meanSCBTSAgent\_Ratio bandit chose noncontextual decisions over contextual ones 10.4\% of the time, SCBTSAgent\_Diff 12.1\% of the time, and meanSCBTSAgent\_Diff 14.2\% of the time. Figure \ref{fig:679-ucb} shows a comparison of the various UCB bandits we evaluated on the same dataset. Again, we observe that all three bandits display the same regret curves. The baseline model, LogisticRegression UCBAgent, made contextual decisions at every timestep, SCBUCBAgent chose the noncontextual decision over the contextual decision a mean percentage of 33.4\% of the times, and meanSCBUCBAgent 14.7\% of the times. 

Figure \ref{fig:1413-ts} shows a comparison of the various Thompson Sampling bandits we evaluated on OpenML dataset id 1413, the Iris dataset. We observe that the four SCB bandits outperform the baseline LogisticRegressionTSAgent. However, it is interesting to note that the number of noncontextual decisions made in favor of contextual ones was minimal. The SCBTSAgent\_Ratio bandit chose noncontextual decisions over contextual ones only .43\% of the time averaged across the 20 runs. Similarly, the meanSCBTSAgent\_Ratio bandit chose noncontextual decisions over contextual ones .43\% of the time, SCBTSAgent\_Diff made a few more noncontextual decisions at 3.9\%, and meanSCBTSAgent\_Diff  at 1.4\%. Figure \ref{fig:1508-ucb} shows a comparison of the various UCB bandits we evaluated on the OpenML dataset id 1508, a user knowledge dataset. Here we observe again all three models performing on par with each other with the SCBUCBAgent choosing noncontextual decision over the contextual decision a mean percentage of 9.6\% and meanSCBUCBAgent 12.1\%. 

The results from the experiments run on publicly available multiclass classification datasets showed us that in many cases, there exists a hybrid policy that reduces dependence on context whilst achieving regret bounds that are on par with fully contextual baseline algorithms. In a few cases, we found a policy that made it possible to outperform a fully contextual policy. Likewise, we also observed a few cases where we were unable to find a delta value that reduced the number of contextual decisions without strictly hurting performance. Overall, we believe that SCB is an algorithm that practitioners can employ if they wish to find policies that reduce the number of the contextual decisions (or conversely, increase the number of common treatments that are assigned) without significantly impacting performance.

\subsection{Experiments on proprietary Dataset}  
In an online video subscription setting, promotional title artwork has undergone a paradigm shift. The burden has shifted from attempting to appeal to as many people as possible with two or three posters to attempting to appeal to a single user in a single online video viewing session. The goal is no longer solely about capturing public joy but about capturing individual attention. Acknowledging that any given title may cover a range of thematic themes with an ensemble of cast members that may each appeal to a particular user in a different way, personalized artwork selection allows online video services to hone in on the varying taste profiles of their member-base. Artwork personalization via online learning with explore exploit and contextual multi-armed bandits is important to the member experience in a video subscription service with thousands of titles and shows available to watch.  The image personalization system algorithmically selects 1 of $N$ possible images from the title image suite to display to a user at a specific point in time or viewing session. The goal is to emphasize different themes through various artwork according to some context (user viewing history, country) in order to capture individual preferences for cast members, genres, artistic themes, etc.  
\paragraph{Evaluation Metrics.} We define the Click through rate (CTR) of an item as the fraction of users who engaged with an item after being presented with it. A controlled randomized uniform exploration of the candidate items is done by a logging policy which provides us with a dataset to evaluate the proposed SCB algorithm.  We compute off-policy replay metrics by following the method described in \cite{Li2011} to evaluate and compare the unbiased offline performance of the various online learning policies we experimented with. This method allows us to answer counterfactual questions based on the logged exploration data. In other words, we can compare offline what would have happened in historical sessions under different scenarios if the recommender system used different algorithms in an unbiased way.
\paragraph{Experimental Set-up.} We compared SCB against a multi-armed contextual bandit policy and a non contextual multi-arm bandit policy. The contextual information of a user is represented as a feature vector provided as input to the model for predicting the probability of reward for each item $i$. Features primarily encapsulate the user’s past engagement behavior. We tested one class of SCB models, based on the same underlying logistic regression model used in production and using the ratio of expected rewards for comparisons against the SCB $\delta$ parameter. If the ratio of the expected reward of the contextual selected arm (or personalized winning image) to the expected reward of the noncontextual selected arm (or unpersonalized winning image) is not above some threshold $\delta$, then we show the unpersonalized winning image. We calculate the policy level take-rate and repeat the experiment for various SCB $\delta$ threshold values. As we sweep the delta from 1 to 10, we move from a fully personalized image selection experience to a fully unpersonalized one, with all members being impressed with the same images for the same titles. We do not anneal the delta rate over the history of contexts, i.e. we keep the delta rate constant. We ran experiments on different asset types and across data streams from different months and days. 

\subsubsection{Compared Policies} \label{compared-policies}
We considered three different policies for selecting the noncontextual winning image. 
\begin{itemize}
\item \textbf{SCB Global Majority Vote} The noncontextual selected image for title $t$ is calculated as the global majority vote image for $t$ across all contexts. Within the context of user $u$ and title $t$, if the ratio between the expected reward of the contextual selected image to the expected reward of the noncontextual selected image is not greater than SCB input parameter $\delta$, then show user $u$ the global majority vote image for $t$. Otherwise, show the member the personalized winning image.
\item \textbf{SCB Country-level Majority Vote} The noncontextual selected image for title $t$ is calculated as the country-level majority vote image for $t$ across all contexts with country location $c$. Within the context of member $m$ with country location $c$ and title $t$, if the ratio between the expected reward of the contextual selected image to the expected reward of the noncontextual selected image is not greater than SCB input parameter $\delta$, then show user $u$ the country-level majority vote image for $t$. Otherwise, show the member the personalized winning image.
\item \textbf{SCB Marketing Default} The noncontextual selected image for title $t$ is set to the marketing default image. Within the context of user $u$ and title $t$, if the ratio between the expected reward of the contextual selected image to the expected reward of the noncontextual selected image is not greater than SCB input parameter $\delta$, then show user $u$ the marketing default image for $t$. Otherwise, show the member the personalized winning image.
\end{itemize}

\subsubsection{Results}
\begin{figure}[t]
    \includegraphics[width=0.5\textwidth]{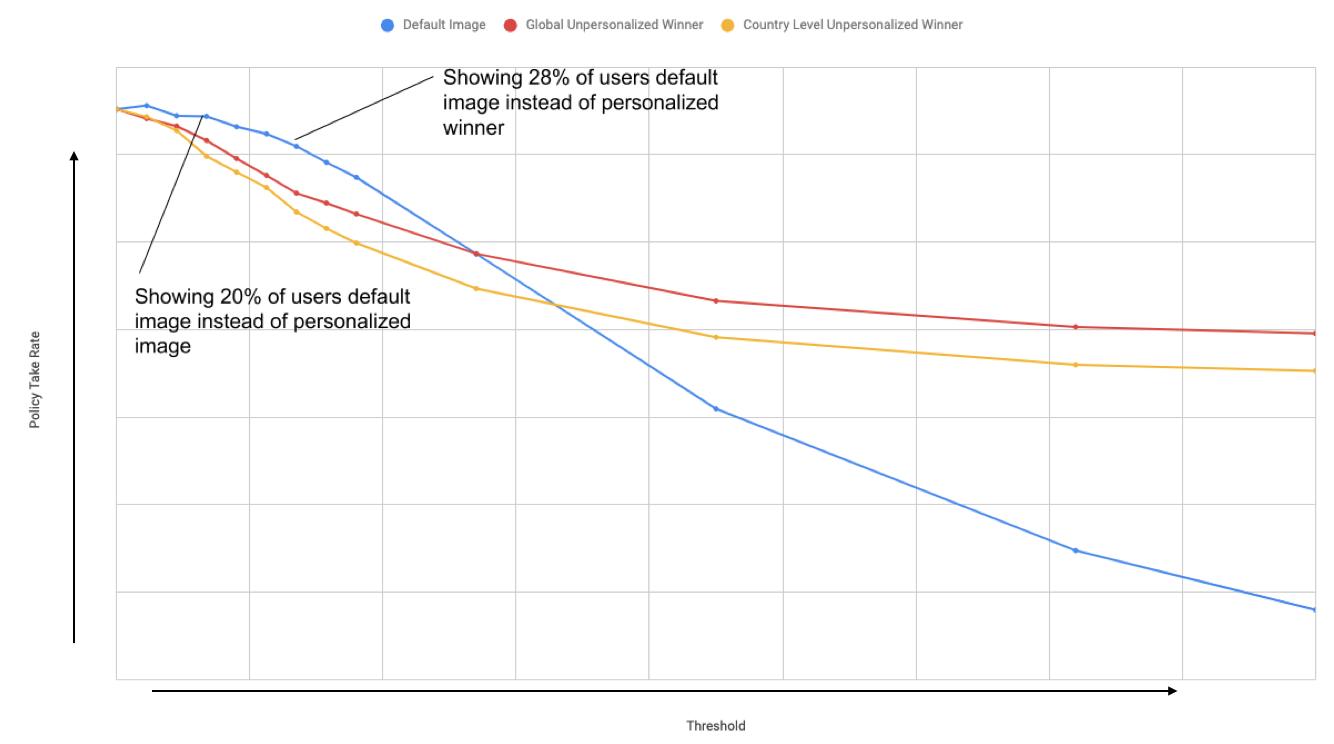}
    \caption{Comparison of three different policies for selecting the noncontextual arm in an industry image personalization system. A policy that fallbacks on the default marketing image during non-contextual SCB decisions performs best until a certain threshold.}
    \label{fig:industry}
\end{figure}
Figure \ref{fig:industry} shows a comparison of the three different policies tested for selecting the noncontextual arm (unpersonalized winning image) on the large-scale proprietary image personalization dataset. Exact details and values are omitted to protect business-sensitive intellectual property. The $x$ coordinate is the SCB threshold value. As we move from left to right, increasing the threshold, we go from a fully personalized experience (threshold of 1) to a fully unpersonalized experienced. That is, we move from a policy that selects the contextual decision every time to one that selects the noncontextual decision every time. The $y$ coordinate is the offline policy take-rate. The higher the value the better. Each of the three lines corresponds to each of the three policies described in Section \ref{compared-policies}. 

The first important observation is that with the SCB Marketing Default policy in particular we are able to increase the number of users who are impressed with the noncontextual decision (in this case, the default image) without a significant negative impact on the overall take-rate. As the figure shows, we are able to show 20\% of the users the noncontextual decision, i.e. default image, instead of the contextual decision, i.e. personalized image, while at the same time achieving a take-rate that is on par with the production policy take-rate that always opts to go with the contextual decision. The second important observation is that at a threshold value that converts 10\% of the contextual decisions to noncontextual ones, we observe an SCB Marketing Default policy take-rate that slightly outperforms the baseline take-rate (the fully contextual policy used in production). In this particular setting, we saw SCB perform best when using the marketing default image as the noncontextual fallback decision. Because as a we increased the SCB threshold value $\delta$, the policy take-rate stayed relatively flat before sharply decreasing. It is worth noting that a policy that shows all users the default image does worse than a fully personalized experience and does worse than a policy that shows all users the global majority vote or country-level majority vote image. Intuitively, this aligns with original motivation to implement image personalization. We conducted the same analysis for different image asset types and across data streams from five different days and saw similar trends. Overall, our offline empirical results on this industry dataset suggest that there exists an SCB policy that reduces dependence on context that can achieve replay take-rates that are on par with or even outperform those achieved by a fully contextual policy.

\section{Conclusion}
In this paper we propose a new family of multi-armed contextual bandits called selectively contextual bandits (SCB) that selectively interpolate between contextual and non-contextual treatment decisions. Using publicly available datasets corresponding to 20 different classification tasks, we have empirically demonstrated that it is possible to increase the number of non-contextual decisions from the policy while achieving similar regret metrics as a fully-contextual policy. In fact, we observe that in some cases, it is possible to even slightly outperform a fully-contextual policy. Further, we demonstrate that SCB is flexible, accommodating different explore/exploit algorithms and allowing the ability to control the amount of contextual decisions that are made using a scheduler to anneal the SCB threshold over the history of time-steps. We hypothesize that an SCB policy is beneficial in creating rich personalized treatments while also increasing the number of shared experiences across users, potentially leading to social participation through network effects. 

\section*{Acknowledgments}
Thank you to Arvind Narayanan, Jeremy Smith, Fernando Amat Gil, Hakan Ceylan, Hossein Taghavi, Christine Doig-Cardet, Sonali Sharma for enabling this body of research and for helpful discussions. This research was supported by the National Defense Science and Engineering Graduate Fellowship and in part by the National Science Foundation under Award CHS-1704444. 

\bibliographystyle{named}
\bibliography{scb}

\begin{thebibliography}{}

\bibitem[\protect\citeauthoryear{Agrawal and Goyal}{2013}]{agrawal2013thompson}
Shipra Agrawal and Navin Goyal.
\newblock Thompson sampling for contextual bandits with linear payoffs.
\newblock In {\em International Conference on Machine Learning}, pages
  127--135, 2013.

\bibitem[\protect\citeauthoryear{Bubeck \bgroup \em et al.\egroup
  }{2012}]{bubeck2012regret}
S{\'e}bastien Bubeck, Nicolo Cesa-Bianchi, et~al.
\newblock Regret analysis of stochastic and nonstochastic multi-armed bandit
  problems.
\newblock {\em Foundations and Trends{\textregistered} in Machine Learning},
  5(1):1--122, 2012.

\bibitem[\protect\citeauthoryear{Chapelle and Li}{2011}]{chapelle2011empirical}
Olivier Chapelle and Lihong Li.
\newblock An empirical evaluation of thompson sampling.
\newblock In {\em Advances in neural information processing systems}, pages
  2249--2257, 2011.

\bibitem[\protect\citeauthoryear{Dimakopoulou \bgroup \em et al.\egroup
  }{2018}]{dimakopoulou2018estimation}
Maria Dimakopoulou, Zhengyuan Zhou, Susan Athey, and Guido Imbens.
\newblock Estimation considerations in contextual bandits.
\newblock {\em arXiv preprint arXiv:1711.07077v4}, 2018.

\bibitem[\protect\citeauthoryear{Dimakopoulou \bgroup \em et al.\egroup
  }{2019}]{dimakopoulou2018balanced}
Maria Dimakopoulou, Zhengyuan Zhou, Susan Athey, and Guido Imbens.
\newblock Balanced linear contextual bandits.
\newblock {\em Thirty-Third AAAI Conference on Artificial Intelligence}, 2019.

\bibitem[\protect\citeauthoryear{Dud{\'{\i}}k \bgroup \em et al.\egroup
  }{2011}]{Dudik2011}
Miroslav Dud{\'{\i}}k, John Langford, and Lihong Li.
\newblock Doubly robust policy evaluation and learning.
\newblock {\em CoRR}, abs/1103.4601, 2011.

\bibitem[\protect\citeauthoryear{Lei \bgroup \em et al.\egroup
  }{2017}]{lei2017actor}
Huitian Lei, Ambuj Tewari, and Susan~A Murphy.
\newblock An actor-critic contextual bandit algorithm for personalized mobile
  health interventions.
\newblock {\em arXiv preprint arXiv:1706.09090}, 2017.

\bibitem[\protect\citeauthoryear{Li \bgroup \em et al.\egroup
  }{2010}]{li2010contextual}
Lihong Li, Wei Chu, John Langford, and Robert~E Schapire.
\newblock A contextual-bandit approach to personalized news article
  recommendation.
\newblock In {\em Proceedings of the 19th international conference on World
  wide web}, pages 661--670. ACM, 2010.

\bibitem[\protect\citeauthoryear{Li \bgroup \em et al.\egroup }{2011}]{Li2011}
Lihong Li, Wei Chu, John Langford, and Xuanhui Wang.
\newblock Unbiased offline evaluation of contextual-bandit-based news article
  recommendation algorithms.
\newblock In {\em Proceedings of the Fourth ACM International Conference on Web
  Search and Data Mining}, WSDM '11, pages 297--306, New York, NY, USA, 2011.
  ACM.

\bibitem[\protect\citeauthoryear{Li \bgroup \em et al.\egroup
  }{2017}]{li2017provably}
Lihong Li, Yu~Lu, and Dengyong Zhou.
\newblock Provably optimal algorithms for generalized linear contextual
  bandits.
\newblock In {\em Proceedings of the 34th International Conference on Machine
  Learning-Volume 70}, pages 2071--2080. JMLR. org, 2017.

\bibitem[\protect\citeauthoryear{Russo \bgroup \em et al.\egroup
  }{2018}]{russo2018tutorial}
Daniel~J Russo, Benjamin Van~Roy, Abbas Kazerouni, Ian Osband, Zheng Wen,
  et~al.
\newblock A tutorial on thompson sampling.
\newblock {\em Foundations and Trends{\textregistered} in Machine Learning},
  11(1):1--96, 2018.

\bibitem[\protect\citeauthoryear{Sutton and
  Barto}{2018}]{sutton2018reinforcement}
Richard~S Sutton and Andrew~G Barto.
\newblock {\em Reinforcement learning: An introduction}.
\newblock MIT press, 2018.

\bibitem[\protect\citeauthoryear{Thompson}{1933}]{thompson1933likelihood}
William~R Thompson.
\newblock On the likelihood that one unknown probability exceeds another in
  view of the evidence of two samples.
\newblock {\em Biometrika}, 25(3/4):285--294, 1933.

\bibitem[\protect\citeauthoryear{Vanschoren \bgroup \em et al.\egroup
  }{2013}]{OpenML2013}
Joaquin Vanschoren, Jan~N. van Rijn, Bernd Bischl, and Luis Torgo.
\newblock Openml: Networked science in machine learning.
\newblock {\em SIGKDD Explorations}, 15(2):49--60, 2013.

\end{thebibliography}

\end{document}